\documentclass[letterpaper, 10 pt, conference]{ieeeconf}  
\IEEEoverridecommandlockouts                              
\usepackage{graphicx}
\usepackage{nohyperref} 
\usepackage{url}  
\usepackage{flushend}
\usepackage{amsmath} 
\usepackage{amssymb}  
\usepackage{enumerate}
\usepackage{subfig}
\usepackage{floatrow}
\usepackage{colortbl}
\usepackage{booktabs}
\usepackage[noadjust]{cite}

\title{\LARGE \bf
Place Categorization and Semantic Mapping on a Mobile Robot 
}
\author{ Niko S\"underhauf, Feras Dayoub, Sean McMahon, Ben Talbot,\\ Ruth Schulz, Peter Corke, Gordon Wyeth, Ben Upcroft, and Michael Milford$^{*}$ 
\thanks{$^{*}$ The authors are with the Australian Centre for Robotic Vision (ACRV), School of Electrical Engineering and Computer Science, Queensland University of Technology, Brisbane,
	Australia. 	email: {\tt\small \{firstname.lastname\}@qut.edu.au}%
}}
\begin{document}
\newcommand{\vect}[1]{\mathbf{ #1}}
\newcommand{\vectg}[1]{{\boldsymbol{ #1}}}
\newcommand{\ggo}{\ensuremath{\mathrm{g^2o}} }
\newcommand{\R}{\mathbb{R}}
\newcommand{\N}{\mathbb{N}}
\newcommand{\Z}{\mathbb{Z}}
\renewcommand{\P}{\mathbb{P}}
\newcommand{\tran}{^\top}
\newcommand{\T}{^\mathsf{T}}
\newcommand{\iT}{^{-\mathsf{T}}}
\newcommand{\inv}{^{-1}}
\newcommand{\func}[2]{\mathtt{#1}\left\{#2\right\}}
\newcommand{\sig}{\operatorname{sig}}
\newcommand{\diag}{\operatorname{diag}}
\newcommand{\argmin}{\operatornamewithlimits{argmin}}
\newcommand{\argmax}{\operatornamewithlimits{argmax}}
\newcommand{\RMSE}{\operatorname{RMSE}}
\newcommand{\RMSEpos}{\operatorname{RMSE}_\text{pos}}
\newcommand{\RMSEori}{\operatorname{RMSE}_\text{ori}}
\newcommand{\RPE}{\operatorname{RPE}}
\newcommand{\RPEpos}{\operatorname{RPE}_\text{pos}}
\newcommand{\RPEori}{\operatorname{RPE}_\text{ori}}
\newcommand{\rpe}{\varepsilon_{\vdelta}}
\newcommand{\achiError}{\bar{e}_{\chi^2}}
\newcommand{\chiError}{e_{\chi^2}}
\newcommand{\normal}[2]{\mathcal{N}\left(#1, #2\right)}
\newcommand{\uniform}[2]{\mathcal{U}\left(#1, #2\right)}
\newcommand{\pfrac}[2]{\frac{\partial #1}{\partial #2}}  
\newcommand{\fracpd}[2]{\frac{\partial #1}{\partial #2}} 
\newcommand{\fracppd}[2]{\frac{\partial^2 #1}{\partial #2^2}}  
\newcommand{\dd}{\mathrm{d}}  
\newcommand{\smd}[2]{\left\| #1 \right\|^2_{#2}}
\newcommand{\E}[1]{\text{\normalfont{E}}\left[ #1 \right]}     
\newcommand{\Cov}[1]{\text{\normalfont{Cov}}\left[ #1 \right]} 
\newcommand{\Var}[1]{\text{\normalfont{Var}}\left[ #1 \right]} 
\newcommand{\Tr}[1]{\text{\normalfont{tr}}\left( #1 \right)}   
\def\sgn{\mathop{\mathrm sgn}}    
\newcommand{\twovector}[2]{\begin{pmatrix} #1 \\ #2 \end{pmatrix}} 
\newcommand{\smalltwovector}[2]{\left(\begin{smallmatrix} #1 \\ #2 \end{smallmatrix}\right)} 
\newcommand{\threevector}[3]{\begin{pmatrix} #1 \\ #2 \\ #3 \end{pmatrix}} 
\newcommand{\fourvector}[4]{\begin{pmatrix} #1 \\ #2 \\ #3 \\ #4 \end{pmatrix}}  
\newcommand{\smallthreevector}[3]{\left(\begin{smallmatrix} #1 \\ #2 \\ #3 \end{smallmatrix}\right)} 
\newcommand{\fourmatrix}[4]{\begin{pmatrix} #1 & #2 \\ #3 & #4 \end{pmatrix}} 
\newcommand{\vA}{\vect{A}}
\newcommand{\vB}{\vect{B}}
\newcommand{\vC}{\vect{C}}
\newcommand{\vD}{\vect{D}}
\newcommand{\vE}{\vect{E}}
\newcommand{\vF}{\vect{F}}
\newcommand{\vG}{\vect{G}}
\newcommand{\vH}{\vect{H}}
\newcommand{\vI}{\vect{I}}
\newcommand{\vJ}{\vect{J}}
\newcommand{\vK}{\vect{K}}
\newcommand{\vL}{\vect{L}}
\newcommand{\vM}{\vect{M}}
\newcommand{\vN}{\vect{N}}
\newcommand{\vO}{\vect{O}}
\newcommand{\vP}{\vect{P}}
\newcommand{\vQ}{\vect{Q}}
\newcommand{\vR}{\vect{R}}
\newcommand{\vS}{\vect{S}}
\newcommand{\vT}{\vect{T}}
\newcommand{\vU}{\vect{U}}
\newcommand{\vV}{\vect{V}}
\newcommand{\vW}{\vect{W}}
\newcommand{\vX}{\vect{X}}
\newcommand{\vY}{\vect{Y}}
\newcommand{\vZ}{\vect{Z}}
\newcommand{\va}{\vect{a}}
\newcommand{\vb}{\vect{b}}
\newcommand{\vc}{\vect{c}}
\newcommand{\vd}{\vect{d}}
\newcommand{\ve}{\vect{e}}
\newcommand{\vf}{\vect{f}}
\newcommand{\vg}{\vect{g}}
\newcommand{\vh}{\vect{h}}
\newcommand{\vi}{\vect{i}}
\newcommand{\vj}{\vect{j}}
\newcommand{\vk}{\vect{k}}
\newcommand{\vl}{\vect{l}}
\newcommand{\vm}{\vect{m}}
\newcommand{\vn}{\vect{n}}
\newcommand{\vo}{\vect{o}}
\newcommand{\vp}{\vect{p}}
\newcommand{\vq}{\vect{q}}
\newcommand{\vr}{\vect{r}}
\newcommand{\vs}{\vect{s}}
\newcommand{\vt}{\vect{t}}
\newcommand{\vu}{\vect{u}}
\newcommand{\vv}{\vect{v}}
\newcommand{\vw}{\vect{w}}
\newcommand{\vx}{\vect{x}}
\newcommand{\vy}{\vect{y}}
\newcommand{\vz}{\vect{z}}
\newcommand{\valpha}{\vectg{\alpha}}
\newcommand{\vbeta}{\vectg{\beta}}
\newcommand{\vgamma}{\vectg{\gamma}}
\newcommand{\vdelta}{\vectg{\delta}}
\newcommand{\vepsilon}{\vectg{\epsilon}}
\newcommand{\vtau}{\vectg{\tau}}
\newcommand{\vmu}{\vectg{\mu}}
\newcommand{\vphi}{\vectg{\phi}}
\newcommand{\vPhi}{\vectg{\Phi}}
\newcommand{\vpi}{\vectg{\pi}}
\newcommand{\vPi}{\vectg{\Pi}}
\newcommand{\vPsi}{\vectg{\Psi}}
\newcommand{\vchi}{\vectg{\chi}}
\newcommand{\vvarphi}{\vectg{\varphi}}
\newcommand{\veta}{\vectg{\eta}}
\newcommand{\viota}{\vectg{\iota}}
\newcommand{\vkappa}{\vectg{\kappa}}
\newcommand{\vlambda}{\vectg{\lambda}}
\newcommand{\vLambda}{\vectg{\Lambda}}
\newcommand{\vnu}{\vectg{\nu}}
\newcommand{\vgo}{\vectg{\o}}
\newcommand{\vvarpi}{\vectg{\varpi}}
\newcommand{\vtheta}{\vectg{\theta}}
\newcommand{\vvartheta}{\vectg{\vartheta}}
\newcommand{\vrho}{\vectg{\rho}}
\newcommand{\vsigma}{\vectg{\sigma}}
\newcommand{\vSigma}{\vectg{\Sigma}}
\newcommand{\vvarsigma}{\vectg{\varsigma}}
\newcommand{\vupsilon}{\vectg{\upsilon}}
\newcommand{\vomega}{\vectg{\omega}}
\newcommand{\vOmega}{\vectg{\Omega}}
\newcommand{\vxi}{\vectg{\xi}}
\newcommand{\vXi}{\vectg{\Xi}}
\newcommand{\vpsi}{\vectg{\psi}}
\newcommand{\vzeta}{\vectg{\zeta}}
\newcommand{\vzero}{\vect{0}}
\newcommand{\cA}{\mathcal{A}}
\newcommand{\cB}{\mathcal{B}}
\newcommand{\cC}{\mathcal{C}}
\newcommand{\cD}{\mathcal{D}}
\newcommand{\cE}{\mathcal{E}}
\newcommand{\cF}{\mathcal{F}}
\newcommand{\cG}{\mathcal{G}}
\newcommand{\cH}{\mathcal{H}}
\newcommand{\cI}{\mathcal{I}}
\newcommand{\cJ}{\mathcal{J}}
\newcommand{\cK}{\mathcal{K}}
\newcommand{\cL}{\mathcal{L}}
\newcommand{\cM}{\mathcal{M}}
\newcommand{\cN}{\mathcal{N}}
\newcommand{\cO}{\mathcal{O}}
\newcommand{\cP}{\mathcal{P}}
\newcommand{\cQ}{\mathcal{Q}}
\newcommand{\cR}{\mathcal{R}}
\newcommand{\cS}{\mathcal{S}}
\newcommand{\cT}{\mathcal{T}}
\newcommand{\cU}{\mathcal{U}}
\newcommand{\cV}{\mathcal{V}}
\newcommand{\cW}{\mathcal{W}}
\newcommand{\cX}{\mathcal{X}}
\newcommand{\cY}{\mathcal{Y}}
\newcommand{\cZ}{\mathcal{Z}}
\newcommand{\fA}{\mathfrak{A}}
\newcommand{\fB}{\mathfrak{B}}
\newcommand{\fC}{\mathfrak{C}}
\newcommand{\fD}{\mathfrak{D}}
\newcommand{\fE}{\mathfrak{E}}
\newcommand{\fF}{\mathfrak{F}}
\newcommand{\fG}{\mathfrak{G}}
\newcommand{\fH}{\mathfrak{H}}
\newcommand{\fI}{\mathfrak{I}}
\newcommand{\fJ}{\mathfrak{J}}
\newcommand{\fK}{\mathfrak{K}}
\newcommand{\fL}{\mathfrak{L}}
\newcommand{\fM}{\mathfrak{M}}
\newcommand{\fN}{\mathfrak{N}}
\newcommand{\fO}{\mathfrak{O}}
\newcommand{\fP}{\mathfrak{P}}
\newcommand{\fQ}{\mathfrak{Q}}
\newcommand{\fR}{\mathfrak{R}}
\newcommand{\fS}{\mathfrak{S}}
\newcommand{\fT}{\mathfrak{T}}
\newcommand{\fU}{\mathfrak{U}}
\newcommand{\fV}{\mathfrak{V}}
\newcommand{\fW}{\mathfrak{W}}
\newcommand{\fX}{\mathfrak{X}}
\newcommand{\fY}{\mathfrak{Y}}
\newcommand{\fZ}{\mathfrak{Z}}

\maketitle
\thispagestyle{empty}
\pagestyle{empty}
\begin{abstract}
In this paper we focus on the challenging problem of place
categorization and semantic mapping on a robot without environment-specific
training. Motivated by their ongoing success in various visual recognition
tasks, we build our system upon a state-of-the-art convolutional network.
We overcome its closed-set limitations 
by complementing the network with a series of one-vs-all
classifiers that can learn to recognize new semantic classes online. 
Prior domain knowledge is incorporated by embedding the classification
system into a Bayesian filter framework that also ensures temporal coherence.
We evaluate the classification accuracy of the system on a robot that maps a variety of places on our
campus in real-time. We show how semantic information can boost robotic object
detection performance and how the semantic map can be used to modulate the
robot's behaviour during navigation tasks. The system is made available
to the community as a ROS module.
\end{abstract}

\section{Introduction}
\begin{figure}[t]
   \centering
    \includegraphics[width=0.9\linewidth]{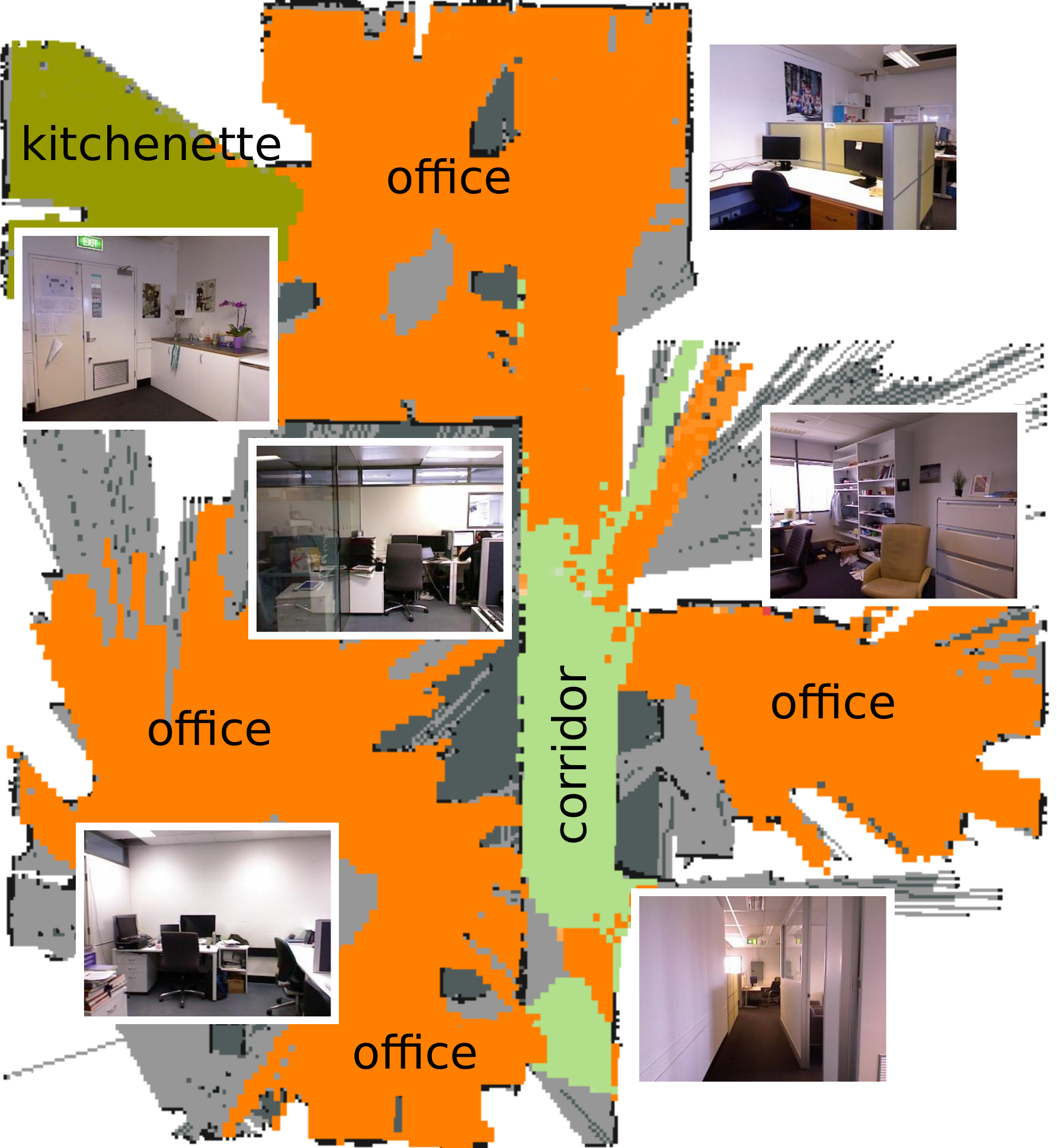}
     \caption{A semantic map generated by the system described in
     this paper. The colors encode the semantic categories of different places
     encountered in the environment. The figure shows a map of an office
     environment (orange) with a kitchenette (dark green) and a long corridor (light green). 
     }
 \label{fig:overallView}
 \end{figure}

To become truly ubiquitous, mobile service robots that operate in human-centered complex indoor and outdoor
environments need to develop an understanding of their surroundings that goes
beyond the ability to avoid obstacles, autonomously navigate, or build maps. 
Truly useful robots need to be able to extract semantic
information about the place they operate in \cite{Galindo08}. Instead of merely answering the
question of ``Where am I?'' \cite{Borenstein96} that often links to the problems
of localization or SLAM, robots should also know ``What is this place like?'' to
aid higher level decision processes, to infer high-level information about an
environment, to ease robot-human interaction and to modulate the robot's
behaviour.
The problem of assigning a semantic label to places or parts of the environment is often referred
to as \emph{place categorization} \cite{Pronobis09, Xiao14}, or -- when combined with
creating a map (see Fig. \ref{fig:overallView}) -- \emph{semantic mapping} \cite{Pronobis12, Hemachandra14}.
To address this challenge we focus on \emph{transferable} and \emph{expandable} semantic
place categorization and mapping for robotics. \emph{Transferable} means the place
categorization does not require environment-specific training.
To achieve this, we leverage the
recent success of convolutional networks (ConvNets) in the computer vision
community where networks were recently trained specifically for the task of
place categorization \cite{Zhou14}.
In contrast to state-of-the-art semantic mapping systems in robotics \cite{Ranganathan10, Wu11,
Pronobis09, Pronobis12} these networks generalize well and do not have to be re-trained or fine-tuned
for specific environments. 
However, ConvNets can only recognize the
classes they have been trained on. This \emph{closed-set} constraint is ubiquitous in
computer vision but poses an important limitation for robotics applications that
aim at long-term autonomous operations and life-long learning. We overcome the
closed-set limitation and present a novel \emph{expandable} classification system by
complementing the ConvNet with one-vs-all classifiers that are computationally cheap to train and can learn to
recognize new classes online.
In typical benchmark applications in computer vision each image is treated individually.
In this paper we exploit the fact that robots see a temporally coherent \emph{stream} of image data and
embed the semantic classifier in a Bayesian filter framework. This allows us to
correct spurious misclassifications and incorporate prior knowledge. We 
demonstrate that more coherent results are achieved when interpreting semantic classification
as a probabilistic estimation problem with first order Markov properties.
In addition to applying a state-of-the-art ConvNet-based classifier on a robot, and showing
that no environment-specific training is required for robotic place
categorization, our paper provides the following contributions:
\begin{enumerate}
   \item We overcome the \emph{closed-set} limitation of the ConvNet classifier by training
     computationally cheap additional one-vs-all classifiers that can learn to
     recognize new classes online.
   \item To benefit from the temporal coherence between consecutive camera
     images, we integrate this dual classification system in a Bayesian filter
     framework. 
   \item We combine the place categorization with a robotic mapping system and
     test it in real-time on a large dataset in a variety of indoor and outdoor
     places. 
  \item We demonstrate that semantic place information can boost \emph{object}
    detection and recognition on a robot.
  \item We demonstrate how a semantic map can be used to modulate robotic
    behaviour in navigation tasks.
  \item We provide the complete system to the community as a ROS module.
\end{enumerate}
To the best of our knowledge such a comprehensive robotic semantic mapping
system has not been proposed before.
In the following we discuss related work in the field in
Section~\ref{sec:related} before introducing our system in
Section~\ref{sec:approach}. Section~\ref{sec:experiment} presents the experimental setup and the
dataset used for evaluation.  Finally we draw conclusions and discuss future
work in Section~\ref{sec:future}.

\section{Related Work}\label{sec:related}
The topic of vision-based semantic scene categorization has been explored both
in the robotics and computer vision community. The SUN (Scene
UNderstanding) challenge initiated by Xiao and Torralba et al. has driven this field of
research forward and resulted in a number of benchmark papers, e.g.
\cite{Xiao10, Xiao14}. Very recently, \cite{Zhou14} achieved a significant
improvement of place semantic categorization on the SUN benchmark by training a
convolutional network for this task.
\subsection{Semantic Mapping in Robotics}
Wu et al. \cite{Wu09} defined the visual place categorization problem for robotics using a purely
appearance-based method based on CENTRIST features \cite{Wu11}. Their system has been
trained on image sequences collected in six different apartment houses and was
able to distinguish between typical semantic room categories like \emph{living
room} or \emph{bathroom}. A similar system has been proposed by
\cite{Ranganathan10} and was tested on the same dataset. In contrast to \cite{Wu09} they use
SIFT features extracted in a dense grid. Although both papers use a
leave-one-out approach for training and testing (i.e. training on data collected
in 5 of the 6 houses, and testing on the 6th), the visual similarity between
the apartments and therefore the similarity between training and testing data is high.
In contrast to the aforementioned purely vision-based methods, Zender et al.
\cite{Zender08} combine laser-based place categorization with 
vision-based object detection and a ``commonsense ontology'' to build a
so called \emph{conceptual map} of the environment that contains semantic
information.  In \cite{Pronobis09}, Pronobis et al. examined another place classification system
that combined sensor data from a camera and a 2D laser range finder to
assign semantic labels like \emph{corridor}, \emph{office} or \emph{meeting
room} in an office building. They combine this with a metric SLAM system to
accumulate class labels over time and generate a semantic map of the
environment. Although the authors separate
their training and testing environments by using different floors of the same
building, the visual similarity between the training and testing instances is  
very high. In later work \cite{Pronobis12}, the same authors extended their system to
include more sensor modalities like object information, or human input in addition to the
visual scene appearance and the geometry information obtained by the laser.
Again they separate training and testing environment by using different floors
in the same building. Unfortunately, their dataset is not publicly available, so
no benchmarking against their results was possible.
Apart from appearance based methods, other authors explored semantic mapping
that relies on the detection of objects to infer semantic information about the
current place (e.g. a cereal box is more likely to be spotted in a kitchen, while
a stapler indicates being in an office). An early example is \cite{Meger08}.
Another research topic that is closely related to semantic mapping is the discovery of
places as recently demonstrated by Murphy and Sibley \cite{Murphy14} as well as
Paul et al. \cite{Paul14, Paul12}. In contrast to the semantic mapping methods
described above, this research focussed on finding meaningful clusters in the
sensor data acquired by a robot over longer periods of time and did not assign
semantic labels to them. 
These \emph{summaries} of perception or experience over time can 
ease the generation of annotated training data over time.
\subsection{Features for Semantic Classification}
A commonality between all of the aforementioned vision-based systems is that
they rely on a fixed set of hand crafted features that
are extracted from the images and then used for classification. For example
\cite{Ranganathan10} uses dense SIFT, \cite{Wu09} uses CENTRIST \cite{Wu11}, 
\cite{Pronobis12} relies on SURF and CRFH and so on.
However, a recent trend in computer vision, and especially in the field of
object recognition and detection, is to exploit \emph{learned} features using
deep convolutional networks (ConvNets).
The most prominent example of this trend is the annual \emph{ImageNet Large
Scale Visual Recognition Challenge} where for the past two years many of the
participants have used ConvNet features \cite{Russakovsky14}.
The concept of convolutional networks is not new and was proposed by
LeCun et al. \cite{LeCun89} to recognize hand-written digits. 
Their popularity has risen ever since algorithmic improvements such as
dropout and rectified linear units~\cite{Hinton12, Dahl13} and the
widespread availability of GPUs to train these models.
Several research groups have shown that ConvNets outperform more classical
approaches for object classification or detection that are based on hand-crafted
features \cite{Krizhevsky12, Sermanet13, Donahue13, Girshick14, Razavian14}.
Recently \cite{Zhou14} used ConvNets to beat all competing approaches on the task of semantic
place categorization.
\subsection{The Closed-Set Limitation}
An important limitation of ConvNets and many other classifiers is the implicitly built in
\emph{closed-set} assumption. The classifier is trained on a fixed set of 
classes and is never presented a new or unknown class during test time.
While this constraint is widespread in many computer vision and machine learning
benchmarks \cite{Scheirer13}, it is not a realistic assumption to make in
long-term robotics applications.
Overcoming this limitation is currently an active field of research in machine
learning and computer vision \cite{Saffari09, Scheirer13, Scheirer14}, but did
not yet converge to a widely accepted solution.
\subsection{Summary}
Despite the large body of work
on semantic mapping and scene classification in the robotics community, most
research systems lack a clear separation between training and testing data which
makes their transferability and generalizabilty questionable. Also, to the best
of our knowledge, the combination of a convolutional network, a set of
computationally cheap one-vs-all classifiers for online learning of new classes,
and a Bayes filter that enforces temporal coherence has not been explored for vision-based semantic mapping
before.

\section{System Overview}
\label{sec:approach}
Our proposed place categorization and semantic mapping system consists of four main
parts:
\begin{enumerate}
  \item a convolutional neural network that classifies each image individually, 
  \item one-vs-all classifiers that recognize scene classes the network was
    \emph{not} trained on,
  \item a Bayesian filter to exploit temporal coherence and remove spurious false
    classifications, and
  \item the mapping subsystem that gradually builds a map using the resulting place
    labels. 
\end{enumerate}
We describe each part in the following, before presenting experiments and
evaluations in the next section.
We provide the complete system as a ROS module for download on our website
http://tinyurl.com/semantic-mapping-QUT.
\subsection{Transferable Place Categorization}
To classify each camera image individually, we leverage the recent successes of
Convolutional Networks for various visual recognition tasks. We use the
\texttt{Places205} network
recently published by Zhou et al. \cite{Zhou14} since it is the state of the art in place
categorization and outperforms all competing methods on various benchmark
datasets\footnote{In an earlier version of this paper we trained a nonlinear SVM
on the \texttt{fc7} layer of the \texttt{AlexNet} provided by \texttt{Caffe}
\cite{Jia14}. This system achieved a top-1 accuracy of 46.2\% on the SUN-397
benchmark, thus achieving state-of-the-art performance, only beaten by
\cite{Sanchez13} at that time which achieved 47.2\%. While this paper was in
preparation, \cite{Zhou14} published their specialized \texttt{Places205}
network which achieves 54.3\%, thus outperforming all previous approaches by a
large margin. We therefore decided to switch to \texttt{Places205}.}.
The \texttt{Places205} network follows the same principled
architecture as \texttt{AlexNet} \cite{Krizhevsky12} but was trained
specifically for the task of place categorization. The training dataset
comprised 2.5 million images of 205 semantic categories, with at least 5,000
images per category. The images originated from several internet sources such as
Google Images, Bing, and Flickr. The images were labeled by human workers using
Amazon Mechanical Turk. The large number and variety of the training dataset
ensures that the resulting classifier generalizes well and does not need to be
retrained or fine-tuned when deployed in environments it has not seen during
training. This ensures that our semantic mapping system is \emph{transferable}
and can be deployed on any robot in a variety of environments.
The input to the \texttt{Places205} ConvNet are RGB images that are resized to $227\times
227\times3$ pixels, independent of their initial aspect ratio. The network's output layer
\texttt{prob} represents the discrete probability distribution
$p(\vx_t|\cI_t)$ over all 205 known classes $x_i$, given the current image
$\cI_t$. The network processes a single image in approximately 30 ms on a Nvidia
Qudaro K4000 GPU which is more than sufficient for typical robotics applications.
\subsection{Expandable Place Categorization}
\label{sec:expand}
A major difference between the computer vision community
and robotics is the \emph{closed set} assumption. Most object detection or
scene classification benchmarks in computer vision assume that all classes  are
known during training, and that the classifier is presented only
images of one of the known classes during testing \cite{Russakovsky14, Zhou14}.
This is called \emph{closed set} classification.
However, research in robotics aims at life-long operations and long-term
autonomy over extended periods of time. Inevitably, the robot will be faced with
scene categories\footnote{The same is true for object recognition
tasks, where the 1000 object classes in ImageNet might not be sufficient or
specialized enough for robotics tasks.} that were not part of the initial training set, but are
important for the robot's mission. Being able to extend the classification
framework with new classes during deployment therefore is crucial.
We show how the place categorization based on the \texttt{Places205}
network can be expanded by a set of new classes $y_i$ that are not part of the original 
training set: We propose to train a \emph{one-vs-all} classifier that
distinguishes the new class $y_i$ from the already known classes $\{x_{0\dots n},
y_{0\dots i-1}\}$.
The advantage of this approach is that it is not necessary to retrain the ConvNet, which would be
computationally expensive (typical training times are in the order of days) and would require
a lot of training images (in the order of hundreds or thousands) of the new
class. In contrast, a Random Forest one-vs-all classifier can be trained in
under a minute using only a few (in the order of 10-100) training images. We
let the classifier use the output of the \texttt{fc7} layer of 
the \texttt{Places205} network as a feature vector. The \texttt{fc7} layer is the last
\emph{generic} (i.e. class independent) fully connected layer in the network.
The layers \texttt{fc8} and \texttt{prob} have 205 output neurons since they are
specifically tailored for the task of recognizing the 205 classes from the
training dataset. 
As mentioned before, $p(\vx_t | \cI_t)$ -- the discrete probability distribution
over $n=205$ class labels $x_i$ -- is the classification result of the \texttt{Places205}
network, given the current image $\cI_t$. Now $p(y_i | \cI_t)$ denotes
the result of one of the one-vs-all classifiers that is trained to classify the
new class $y_i$. Let
\begin{equation}
  \hat\vx = \left(x_0, x_1, \dots x_n, y_0, \dots, y_m \right)
  \label{}
\end{equation}
denote the \emph{combined} vector of class labels. Then we define the combined
likelihood $\cL(\cI_t | \hat\vx_t)$ as
\begin{equation}
  \cL(\cI_t | \hat\vx_t) = \left(p(x_0|\cI_t), \dots, p(x_n|\cI_t),
  p(y_0|\cI_t),\dots, p(y_m|\cI_t)\right)
  \label{eq:likelihood}
\end{equation}
Re-normalization distributes the probability between the $n$ classes known to the ConvNet
classifier and the $m$ additional classes known to the one-vs-all classifiers in a natural
way. Notice that this assumes independence between the class labels $x_{0\dots
n}$ and $y_{0\dots m}$ as well as pairwise independence between any $y_i$ and $y_j$. 
\subsection{Bayesian Filtering over Class Labels for Temporal Coherence}
\label{sec:bayes}
Typical computer vision benchmarks for place categorization or object detection treat each
image individually \cite{Russakovsky14, Zhou14}. In contrast, most of the
observed sensor data in robotics have a \emph{temporal} dimension. Knowing that two images were
observed consecutively can be a strong source of information that can be
exploited by using Bayesian filtering techniques.
We interpret the robotic place categorization problem as a
\emph{probabilistic estimation} problem and
estimate the discrete distribution $p(\hat\vx_t | \cI_{0:t})$ over all possible place
labels $x_i$, given all the observed images $\cI_{0:t}$ from the past until now.
Assuming first order Markov properties, this leads to the following
well-known Bayesian filter step:
\begin{equation}
  p(\hat\vx_t | \cI_t) = \cL(\cI_t|\hat\vx_t) \cdot p(\hat\vx_{t-1} | \cI_{t-1})
  \label{eq:bayes}
\end{equation}
where $\cL(\cI_t|\hat\vx_t)$ is the combined likelihood defined in
(\ref{eq:likelihood}).
\subsection{Incorporating prior knowledge}
\label{sec:prior}
Interpreting place categorization as a Bayesian estimation problem allows us to
incorporate other sources of information in a very natural way. For instance we
might know that many of the 205 categories \texttt{Places205} can recognize are
unlikely or even impossible to be observed in the environment the robot is
deployed in. This kind of knowledge $p(\hat\vx)$ can be easily incorporated
by an additional prior term:
\begin{equation}
  p(\hat\vx_t | \cI_t) = p(\hat\vx) \cdot \cL(\cI_t|\hat\vx_t) \cdot p(\hat\vx_{t-1} | \cI_{t-1})
  \label{eq:bayes_prior}
\end{equation}
\subsection{Semantic mapping}
\begin{figure}[tp]
	\begin{center}
		\includegraphics[width=2in]{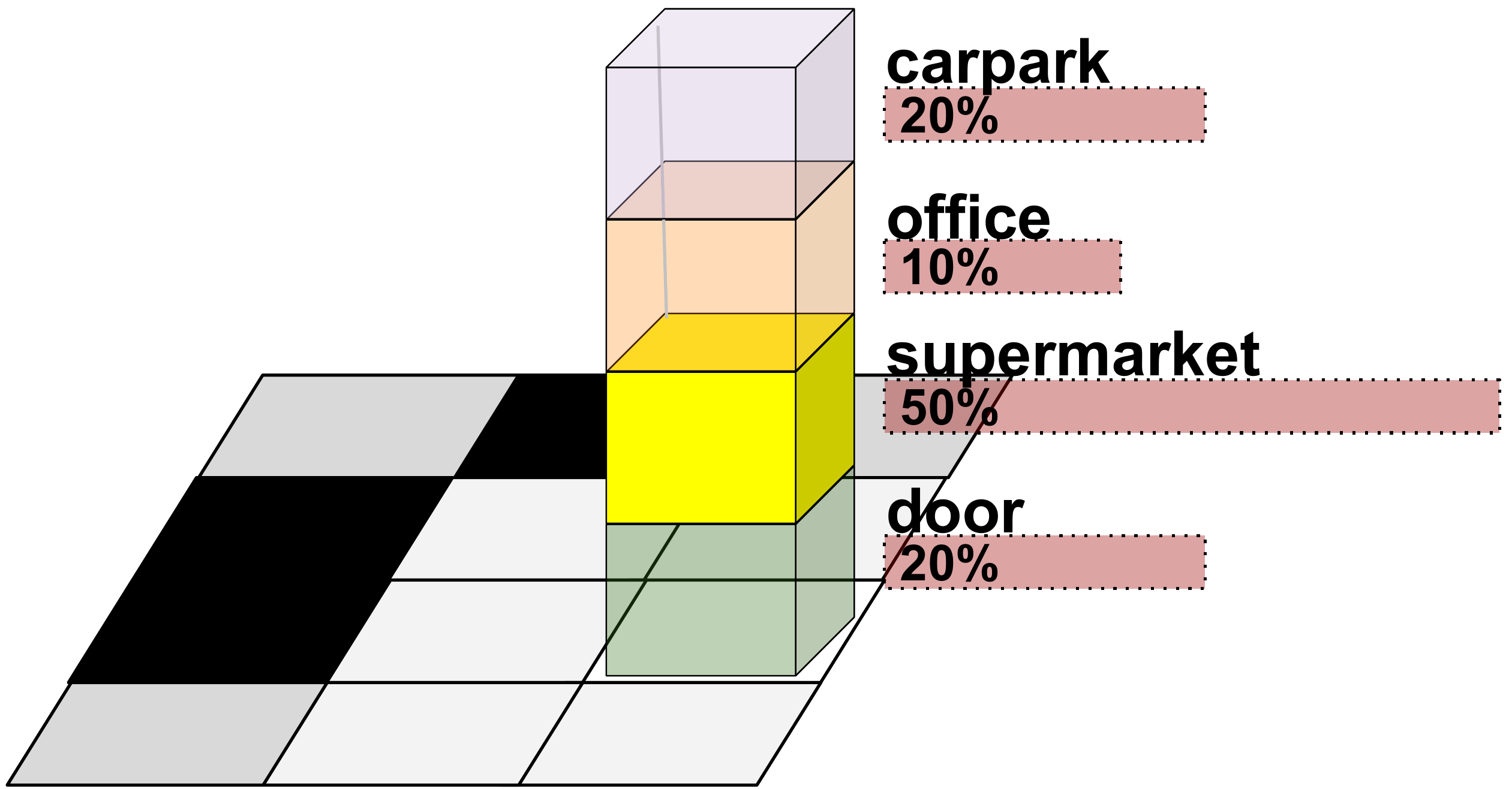}  
	\end{center}
	\caption{An illustration of the semantic map structure.
        Level zero of the map is an regular occupancy grid but the higher levels
        encode the probabilities that a grid cell belongs to a certain semantic
        category. Each layer in the map represents one semantic category.
        The cells are updated according to equation~(\ref{eq:2}).                
        }
	\label{fig:semMap}
\end{figure}

The place categorization component described in the previous section creates a
probability distribution $p(\hat\vx_i|\cI_t)$ over the known class labels, given the current camera
image $\cI_t$. This continuous stream of classification results is the input to
the semantic mapping component, along with a laser range scan which aids the map building. 
To build a combined semantic and metric map of the environment, we apply the
occupancy grid mapping algorithm and maintain one map layer per semantic
category. Fig. \ref{fig:semMap} illustrates this concept. Instead of expressing
the probability of being occupied or free, each cell in these semantic layers
expresses the probability of belonging to a certain semantic category.
This is achieved by propagating the class probability $P(\vl_i|z_t)$ along the laser
rays that are within the field of view of the camera and updating the penetrated
map cells using the usual recursive Bayes filter update method \cite{Elfes89} for occupancy
maps. This way, all unoccupied cells that are within 5 meters of the robot's
position and within the camera's field of view are
updated with the currently observed semantic label. 
The probability that a cell $k$ is representing a place category $\hat\vx_i$ given the classification result $z_{1:t}$ are : 
\begin{multline}
\label{eq:1}
           p_{k}(\hat\vx_i|\cI_{1:t}) =\\ \lbrack 1+
                                          \frac{1-
                                          p_k(\hat\vx_i|\cI_t)}{p(\hat\vx_i|\cI_t)} 
                                          \frac{1-
                                          p_k(\hat\vx_i|\cI_{1:t-1})}{p(\hat\vx_i|\cI_{1:t-1})}
			                  \frac{p(\hat\vx_i)}{1 - p(\hat\vx_i)}
                                 \rbrack^{-1}                    
\end{multline}
where $p_k(\hat\vx_i|\cI_t)$ is the probability that cell $k$ is of place category
$\hat\vx_i$ given the currently observed image $\cI_t$, while $p_k(\hat\vx_i|\cI_{t-1})$ is
the previous estimate and $p(\hat\vx_i)$ is a prior probability.
For better performance, our implementation uses a log-odds representation,
which results in the following simple update equation \cite{Hornung13}:
\begin{equation}
\label{eq:2}
           \cL_{k}(\hat\vx_i|\cI_{1:t}) = \cL(\hat\vx_i|\cI_{1:t-1}) +
           \cL(\hat\vx_i|\cI_{t})
\end{equation}
where $\cL(\hat\vx_i) = \log{\frac{p(\hat\vx_i)}{1-p(\hat\vx_i)}}$. 
Finally, a clamping step ensures that $ l_\text{min} \le \cL_{k}(\hat\vx_i|z_{1:t}) \le
l_\text{max}$.
Notice that spurious
false classifications do not permanently corrupt the map since the probabilities
within the cells are adapted gracefully and can be corrected with later
observations.
The resulting map can be used in a variety of tasks. For instance for path
planning different traversal costs can be assigned to different labels to make
the robot avoid busy places during certain times of the day (e.g. the food court
during lunch time).

\section{Experiments, Evaluation and Results}\label{sec:experiment}
\subsection{Place Categorization on a Real Robot}
\begin{figure}[t]
	\begin{center}
		\includegraphics[width=\textwidth]{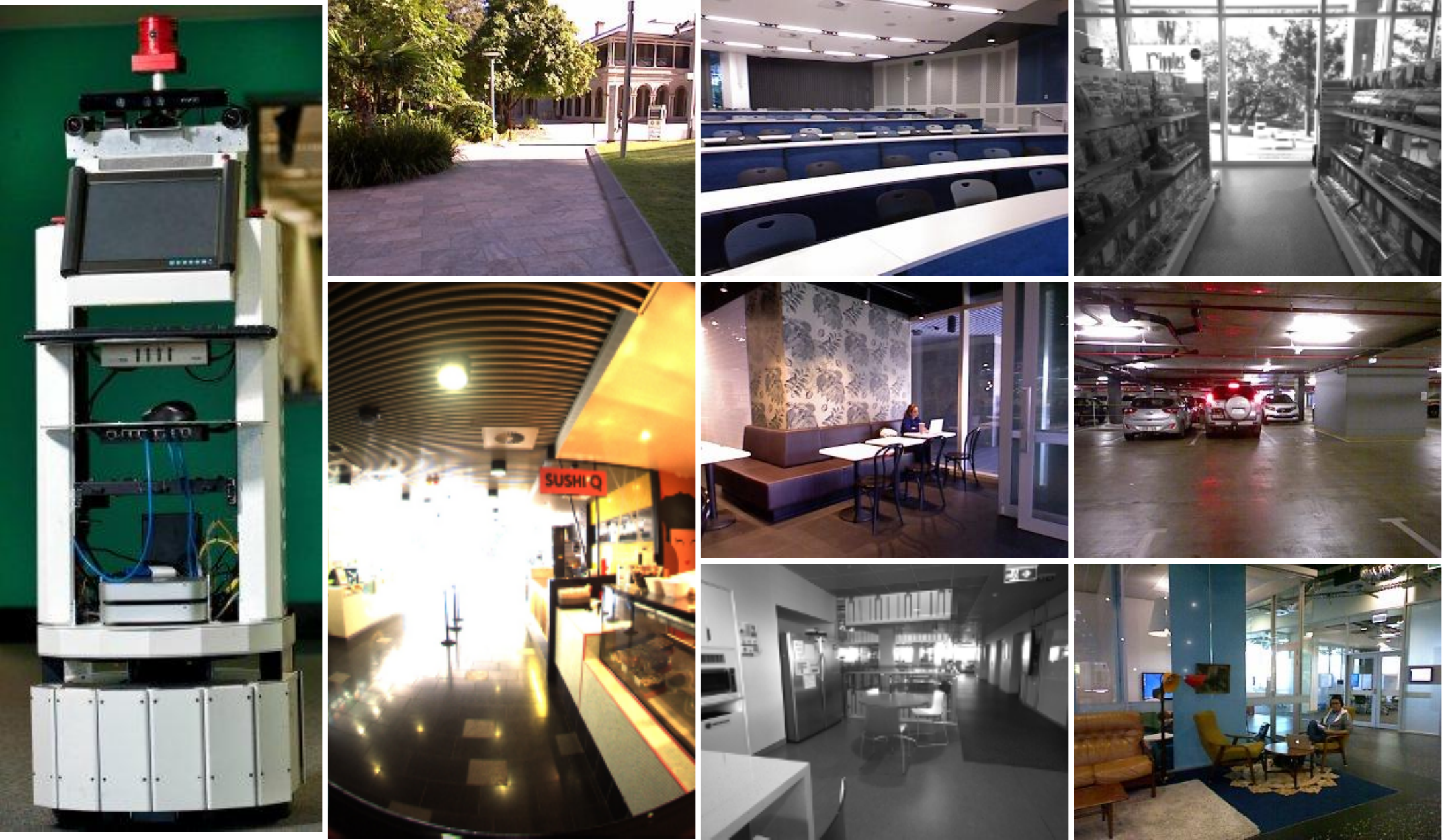}
	\end{center}
	\caption{The Guiabot robot used to evaluate the system and 
		example images from all three cameras in a variety of places:
		Kinect RGB (color), Grayscale, and Ladybug (portrait format). Notice that
		all images are resized to a fixed size of $231\times231$ before
		calculating the ConvNet features. While the change in aspect for the 
		RGB and Grayscale images is minor, the Ladybug image gets squeezed significantly.
		Also notice the low quality of the Ladybug image.}
	\label{fig:guiabot}
\end{figure}

We demonstrate and evaluate the place categorization performance of our proposed
system on a real robot on our university's campus. 
We use the MobileRobots Research GuiaBot shown in Figure~\ref{fig:guiabot} and
test the system on the images of three types of cameras that are mounted on the
robot: 1) the RGB camera from
the Microsoft Kinect (version 1) sensor, 2) Point Grey Grasshopper monochrome
camera and 3) the front facing camera of the Ladybug2, a spherical camera.
\begin{table}[tb]
	\begin{center}
		\begin{tabular}{@{}llll@{}}	
			\toprule
			{\bf Environment}   & {\bf RGB}  & {\bf
                        Grayscale} & {\bf Ladybug} \\ \midrule
        corridor & 100.0\% &  98.4\% &  98.2\% \\ 
	office S11 & 94.4\% &  96.4\% &  97.0\% \\ 
	parking garage & 90.8\% &  86.0\% &  98.0\% \\ 
	foodcourt & 84.2\% &  65.6\% &  48.2\% \\ 
	cafe outside & 66.6\% &  62.2\% &  53.1\% \\ 
	shop & 53.3\% &  56.3\% &  45.9\% \\ 
	lobby & 49.1\% &  \;\,\,--- &  31.2\% \\ 
	lecture theater & 44.5\% &  38.0\% &  35.2\% \\ 
	outdoor & 33.3\% &  3.3\% &  55.9\% \\ 
        \midrule            
	weighted average & 67.7\% & 61.6\% & 59.8\% \\
	\bottomrule
        \end{tabular}
	\end{center}
	\caption{Accuracies for the different cameras in the QUT Gardens Point
        Campus dataset. The bottom row gives the average accuracy weighted by
        the number of recorded frames for each environment. Notice that there
        were no grayscale images recorded for the lobby environment.}
\label{tbl:accuracies}
\end{table}
The test dataset was collected by teleoperating the robot across nine different and
versatile environments on our campus and recording the images from all three cameras as
well as laser scans and odometry. The traversed environments are listed in Table
\ref{tbl:accuracies}.
Since we know the robot is operating on the campus, only the following set of
semantic classes could possibly be encountered: $\hat X=$ \{corridor, classroom, office, parking\_lot,
restaurant, food\_court, kitchen, kitchenette, lobby, supermarket, clothing\_store,
botanical\_garden, coffee\_shop\}. We incorporated this prior information as described in
Section \ref{sec:prior}. The prior probabilities of all classes \emph{not} in
$\hat X$ were set to zero.
We measure the top-1 classification accuracy for each of the nine environments
separately for the images captured by the three different types of cameras
on the robot. The classifier results were checked by a human expert and the
resulting accuracies are summarized in Table \ref{tbl:accuracies}. 
The RGB camera (from the Kinect 1 sensor) produces the best
results, presumably because the camera characteristics are more similar to the
cameras the training set was captured with. Our system 
performed well with grayscale images, with average accuracy 6.1\% below that of
the RGB camera and only performing much worse on the outdoor dataset. This
indicates that color is not an important cue for scene classification in many indoor
environments. The ladybug camera performed worst. We account that to the extreme
deformations that occur (see Fig. \ref{fig:guiabot}) when squeezing the
Ladybug's upright format images into a squared input image for
\texttt{Places205}. 
\subsection{The Effect of the Bayes Filter}
\begin{figure}[tp]
	\begin{center}
		\includegraphics[width=\linewidth]{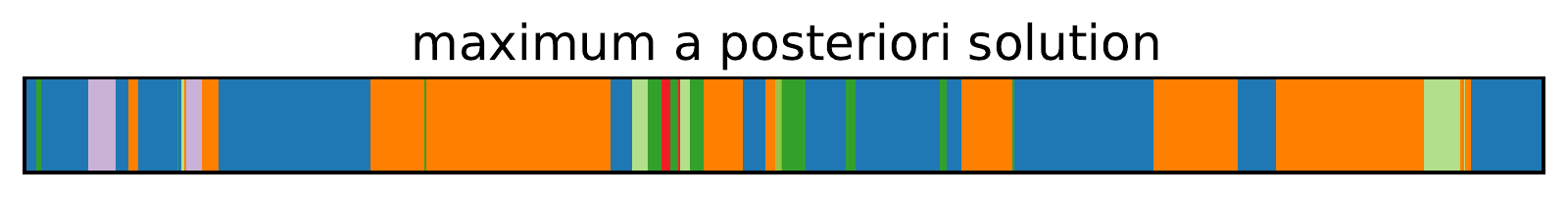}  
                \includegraphics[width=\linewidth]{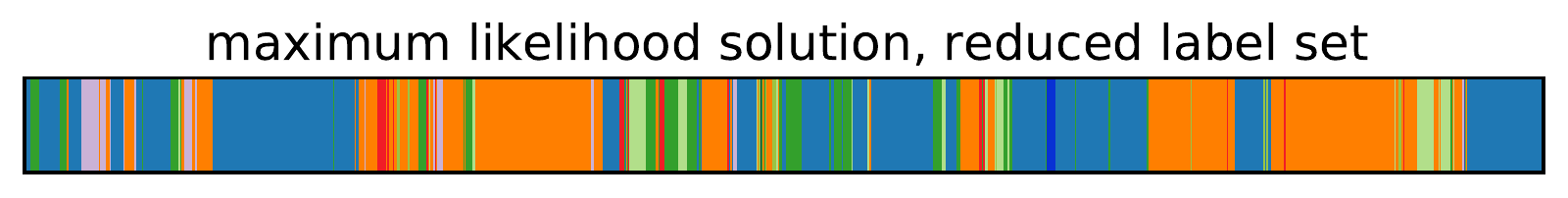}  
                \includegraphics[width=\linewidth]{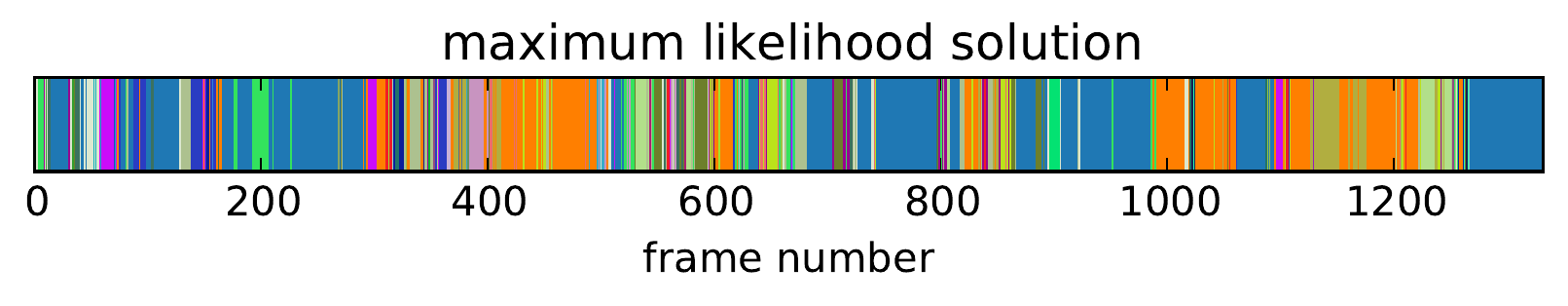}  
	\end{center}
	\caption{Comparing maximum a posteriori and maximum likelihood solutions on an
        indoor dataset. Each color corresponds to a different class
        label. The plots show that the Bayes filter (top) smoothes out spurious
        classification results that would occur when trusting the maximum
        likelihood solution (middle). The bottom plot shows the highly
        fluctuating ML solution without incorporating prior domain knowledge
        (i.e. reducing the prior probability of observing certain scene
        classes).
        }
	\label{fig:bayes:results}
\end{figure}
Fig. \ref{fig:bayes:results} illustrates the positive effects of the Bayes
filter that enforces temporal coherence and incorporates prior knowledge.
The maximum a posterior solution is much more stable then the maximum likelihood solution
alone. Spurious results are smoothed out, reducing false classifications. The
bottom plot shows the maximum likelihood solution when not incorporating prior
domain knowledge, and using a uniform prior probability over all 205 class
labels instead. 
\subsection{Semantic Mapping Results}
\begin{figure}[t]
	\begin{center}
		\includegraphics[width=\linewidth]{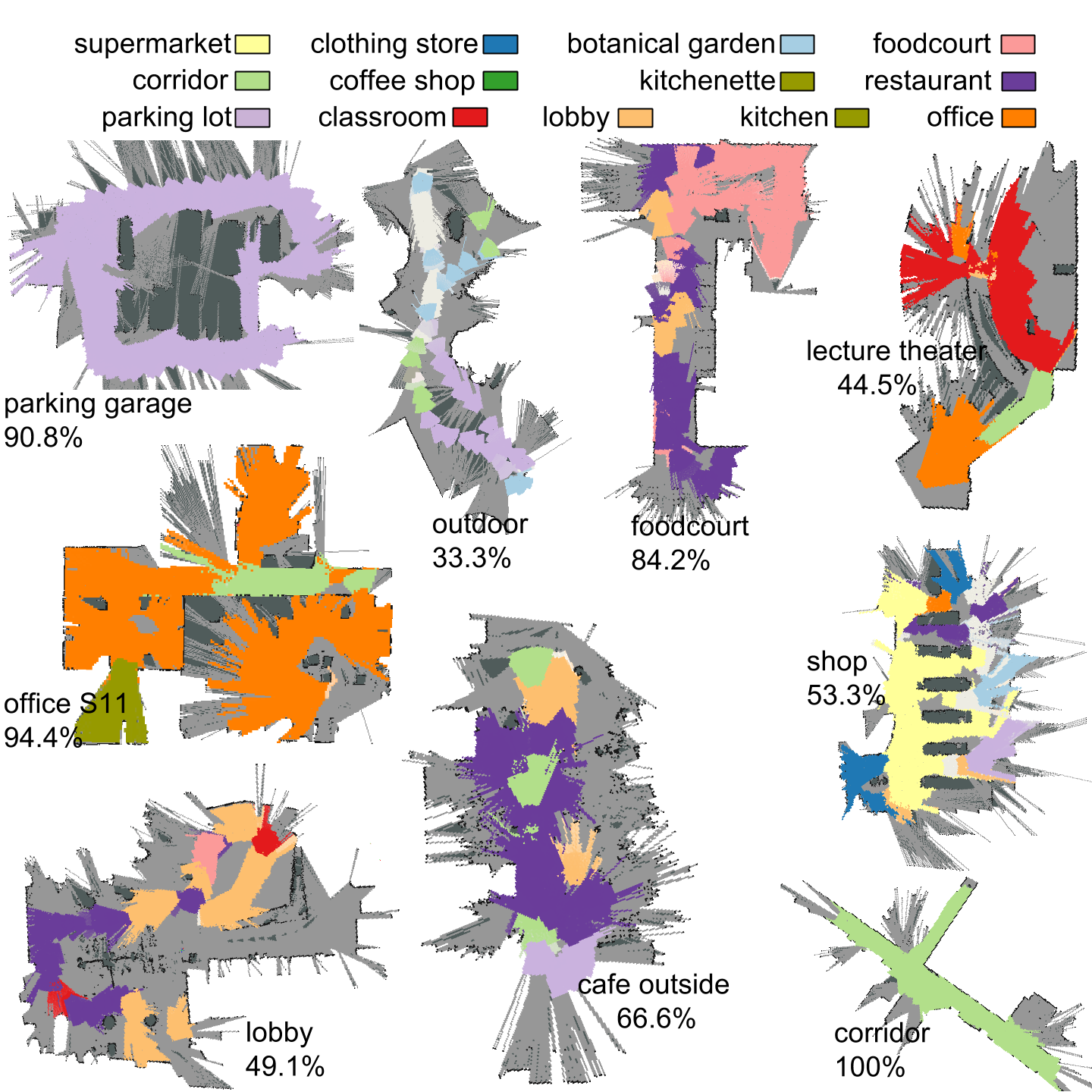} 
	\end{center}
	\caption{Maps created by the semantic mapping system in nine different
        parts of our campus (not drawn to scale). The percentages refer to the
        fraction of correctly classified \emph{images}, not to the correctly
        labeled map area. See the text for further explanations.
          } 
	\label{fig:all_maps}
\end{figure}

Fig.~\ref{fig:all_maps} shows the output of our semantic mapping system using
the RGB images in the nine different tested environments on campus. The map are
color coded where the only the winning labels are rendered. 
The percentage values given in the figure correspond to the fraction of
correctly classified images listed in Table \ref{tbl:accuracies}, but do not
necessarily reflect the correctly classified map area.
Each place has a main category label that describes the place in general,
however, all the nine places contain mixed categories of places. For example,
the office environment, as shown in the left middle in Fig.~\ref{fig:all_maps}
contains actual offices (orange) that are connected by a long corridor (light
green) and a kitchen area (brown). Similarly, not all areas in the
supermarket environment have been assigned the label \texttt{supermarket}
(yellow). A large area in the lower right can be seen colored in dark blue,
representing a \texttt{clothing store}. Manual inspection confirmed this
classification result, since the store actually sells clothes and has them on
display in this part of the shop. 
Towards the top and the right of the supermarket map, large windows lead
out to the open campus. The system created incoherent classifications and
assigned the labels botanical garden and parking lot.
Another particular challenging map is that of the food court.
The classifier correctly assigned the label the \texttt{restaurant} (purple) in areas
with tables and chairs and \texttt{food\_court} (pink) when the robot faces
the actual food stalls. The transition areas have been labeled as \texttt{lobby}.
Given that our test dataset contained many very challenging and cluttered
scenes, we are very content with the classification and mapping results produced
by our system. 
\subsection{Expanding the Classifier with New Classes}
As discussed in Section \ref{sec:expand}, the ability of adding new classes to
the classification system is crucial for long-term operation.
We expanded our system by adding a new class \texttt{door} that
\texttt{Places205} cannot detect. 
We randomly selected 80 positive examples from the
\texttt{elevator-door}\footnote{There is no proper \texttt{door} category in the
dataset.} category of the \texttt{SUN-397} dataset \cite{Xiao10}, and 26,000 negative
images from other categories. Training the classifier on a standard desktop
machine using Python's \texttt{scikit-learn} implementation takes only 57 
seconds on a desktop machine. 
Although the one-vs-all classifier was trained on elevator doors, it 
reliably detects the doors that are typically found in offices and
corridors and so on. We tested this on another dataset (1332 images) we
collected in our office environment, where the robot was driven through multiple
offices and corridors. The system detected 8 of the 10 doors in the environment
and produced two false positive detections. The two false negatives can be
accounted to rapid camera motion (the door was only visible for 4 frames) and
extremely bad lighting conditions (the door was at the end of a long unlit
corridor). For one of the two false positive detections the classifier responded
to a wooden plank on the wall. 
\section{Case Studies: Applications of Semantic Information in a Robotics
Context}
\label{sec:caseStudies}
\subsection{Semantic Place Information Boosts Object Detection}
\begin{figure}[t]
	\begin{center}
		\includegraphics[width=\linewidth]{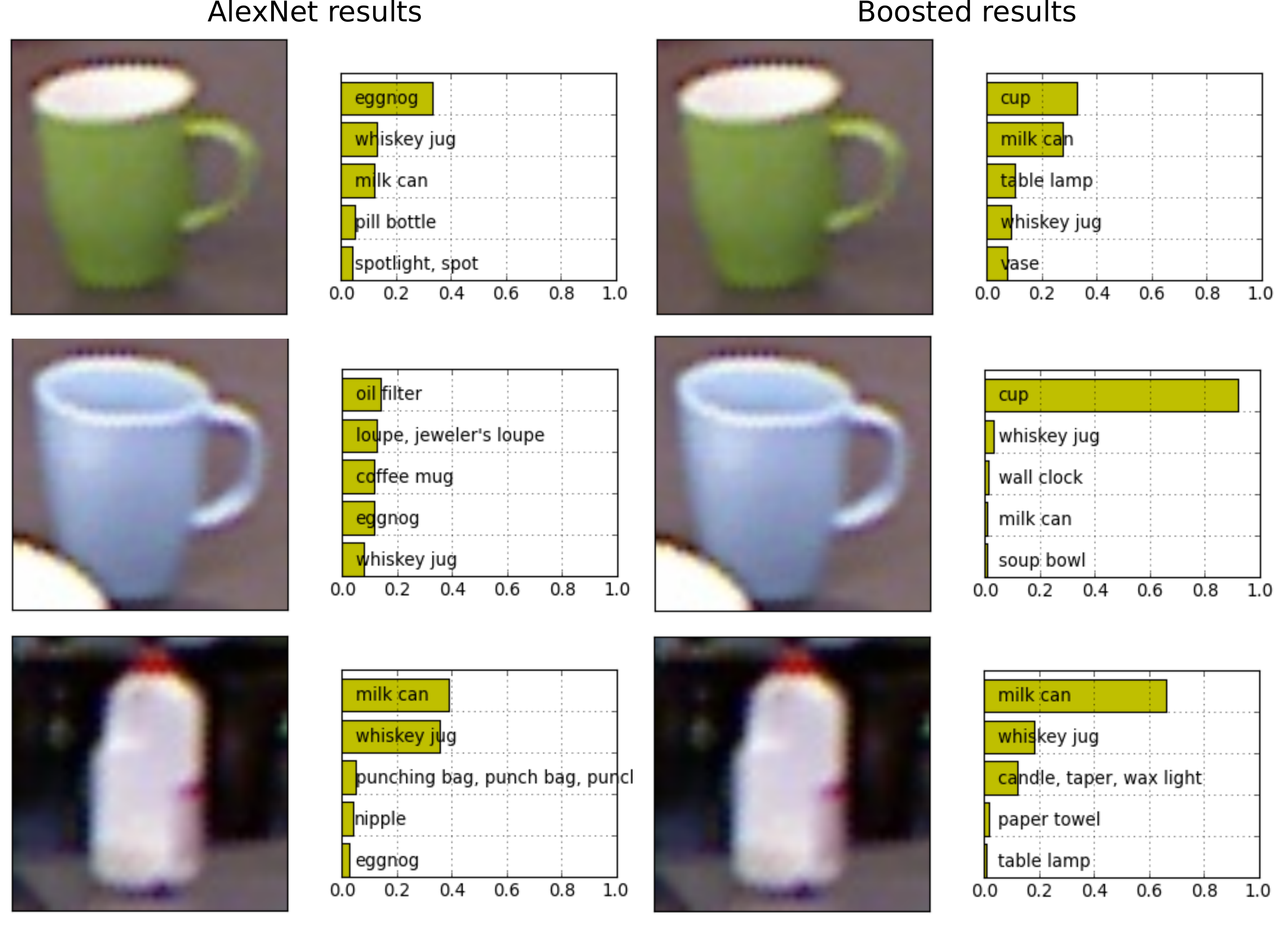} 
	\end{center}
	\caption{Case study I: Semantic place information significantly boosts object
        recognition performances. The examples
        illustrate how the top-5 classification results by \texttt{AlexNet} on
        objects in a kitchen scene can be improved by incorporating prior
        knowledge that is conditioned on the current semantic scene class. Left:
        original results, right: results when 
        exploiting the semantic mapping system and boosting the object classifier with
        prior probability $p(c_i | \hat x^*)$. See text for details.
        }
	\label{fig:objects}
\end{figure}

The semantics of a place provide valuable information that can
boost the performance of object detection and recognition on a robot. We demonstrate this by
running an object detection pipeline inspired by \cite{Girshick14} that consists
of an object proposal step (using EdgeBoxes \cite{Zitnick14}) and a ConvNet classifier
(\texttt{AlexNet} \cite{Krizhevsky12} as imlemented in \texttt{Caffe} \cite{Jia14}).
Similar to \texttt{Places205}, \texttt{AlexNet} provides a discrete probability
distribution over the 1000 object classes it was trained on. We denote this as $p(\vc |
\pi)$, where $\pi$ is the image patch the classifier is applied on. Depending on the
semantic context, different object classes are more
likely to be observed than others. E.g. in a kitchen we expect to see cups and
mugs, but not a motor bike. We propose to exploit such knowledge in a naive
Bayes classifier:
\begin{equation}
  p(\vc|\pi, \hat x^*) \propto p(\vc | \pi)\cdot p(\vc | \hat x^*)
  \label{eq:bayes_objects}
\end{equation}
where $\hat x^*$ is the maximum-a-posteriori solution of the semantic place
category for the currently observed scene and we assumed independence between
$\pi$ and $\hat x^*$ and uniform prior probabilities.
The terms $p(c_i | \hat x^*)$ express the likelihood to observe an object of
class $c_i$ in a scene with label $\hat x^*$. We learned these priors from the
NYU2 dataset \cite{Silberman12} by analysing the ground truth labels and building statistics over
the relative occurrences of object classes for every scene type. Non-occurring
object classes or classes that do not appear in both the NYU2 and the ImageNet
datasets were given a small but non-zero default prior probability.
We tested the combination of our semantic place categorization and object detection
on a robot in a kitchen environment. We found the robot was more
accurate in its object classifications when it had the additional semantic
information available and could apply equation (\ref{eq:bayes_objects}). Fusing both
streams of information increased the top-1 accuracy of correctly detected objects
from 0 \% to 54 \% and the top-5 accuracy from 15\% to 100\% in this experiment.
Examples of the boosted object detection results are shown in Fig.
\ref{fig:objects}.
\subsection{Path Planning on a Semantic Map}
\begin{figure}[t]
	\begin{center}
		\includegraphics[width=0.9\linewidth]{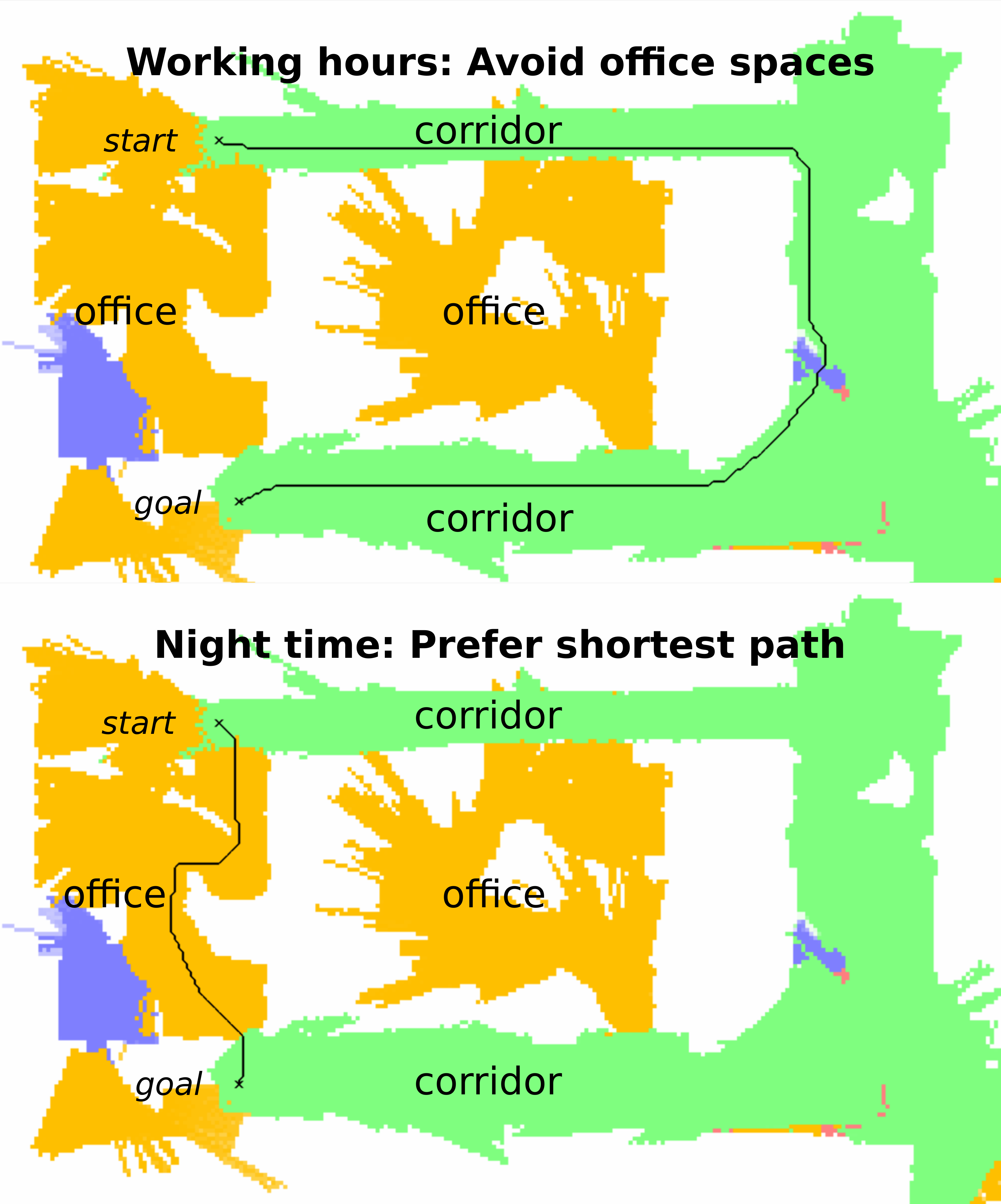} 
	\end{center}
	\caption{Case study II: Path planning on a semantic map. Semantic information can be
        used to modulate the robot's behaviour by translating place labels to
        different cost values for the path planner. This way the robot avoids
        office spaces during working hours (top) but follows the shortest path
        during the night (bottom). 
        } 
	\label{fig:planner}
\end{figure}

We demonstrate how the semantic map created by our system can modulate the behaviour of a robot in an
indoor navigation scenario. In our example a robot has to navigate from its
current position to another place in a workplace environment with offices and
corridors. During working hours the robot should try to 
avoid disturbing humans in their office spaces and rather take longer detours
through the corridors. During night times, the shortest path is always
preferable. 
Such scenarios can be easily implemented when performing the path planning in
the semantic map. Different class labels in the map can be assigned
different \emph{cost} values that are used by a path planner. Fig.
\ref{fig:planner} compares the results of an $\text{A}^*$ path planner when
avoiding offices and when preferring the shortest path.

\section{The Provided ROS Module}
We provide the semantic place categorization system and the semantic mapper as a
ROS module to the community. 
The module consists of two nodes. One node subscribes to an image topic and interfaces
the \texttt{Caffe} framework \cite{Jia14} that implements the \texttt{Places205}
ConvNet. 
The network provides a probability distribution over the 205 known scene types.
On a Nvidia Quadro K4000 GPU, this operation takes 0.031 seconds  for a $320\times 240$ image.  
During the classification process, features from layer \texttt{fc7} are
extracted and passed through the one-vs-all classifiers to detect additional
classes. This step requires additional 3 ms per classifier using a Python
implementation of a random forest classifier on a desktop machine.
After passing the combined likelihoods through the Bayes filter, the posterior
probability distribution is published as a ROS topic. 
A second node subscribes to this topic, the data of a laser range finder, the
robot pose estimate and a grid map created by a SLAM system such as gmapping.
Our node fuses all these information and creates a 3-dimensional map structure
(based on OctoMap) where each layer contains a probability map for a specific semantic
class.
The complete system can be dowloaded from http://tinyurl.com/semantic-mapping-QUT, where additional
information and documentation can be found. This ROS package provides a readily
usable semantic categorization and mapping system to the community that can be
deployed without environment-specific training.
\section{Conclusion and Future work}\label{sec:future}
Our paper introduced a novel transferable and expandable place
categorization and semantic mapping system that requires no environment-specific
training, can be expanded with new classes, and is embedded in a Bayesian filter
framework that allows the incorporation of prior information and enforces
temporal coherence of the classification results.
Semantic information about the environment is an important enabler of more
advanced robotic tasks, especially for human-robot collaboration. Humans
describe places, goals, and objects using semantic categories and it is natural
for them to formulate tasks using these categories. We demonstrated how semantic
information can influence robotic navigation tasks in a workplace, making robot
operations more compliant with human needs.
A second case study demonstrated how semantic information supports robotic
object detection and increases the performance of this equally important visual
recognition problem.
In future work we will extend the system to support  multi-scale or sub-scene
categorization, i.e. assigning labels to parts of the scene. We will also
explore how semantic mapping can guide visual place recognition by partitioning
the search space to similar semantic places and apply the system to various
robotic tasks in the real world.

\bibliographystyle{IEEEtran}
\bibliography{bibfile.bib}
\end{document}